\documentclass[review, authoryear]{elsarticle}

\usepackage{geometry}
\usepackage{lineno}
\usepackage{graphicx}
\usepackage{diagbox}
\usepackage{multirow}
\usepackage{multicol}
\usepackage{amsmath,amsfonts,amssymb}
\usepackage{booktabs}
\usepackage{siunitx}
\usepackage{subfigure}
\usepackage{amsmath}
\usepackage[ruled,vlined]{algorithm2e}
\usepackage[T1]{fontenc}
\usepackage{url}
\usepackage{csquotes}
\usepackage{gensymb}

\begin{document}

\begin{frontmatter}

\title{In-situ animal behavior classification using knowledge distillation and fixed-point quantization}

\author[1]{Reza~Arablouei\corref{cor1}}
\author[2]{Liang~Wang}
\author[1]{Caitlin~Phillips}
\author[1]{Lachlan~Currie}
\author[1]{Jordan~Yates}
\author[2]{Greg~Bishop-Hurley}
\cortext[cor1]{Corresponding author}
\address[1]{Data61, CSIRO, Pullenvale QLD 4069, Australia}
\address[2]{Agriculture and Food, CSIRO, St Lucia QLD 4067, Australia}

\begin{abstract}

We explore the use of knowledge distillation (KD) for learning compact and accurate models that enable classification of animal behavior from accelerometry data on wearable devices. To this end, we take a deep and complex convolutional neural network, known as residual neural network (ResNet), as the teacher model. ResNet is specifically designed for multivariate time-series classification. We use ResNet to distill the knowledge of animal behavior classification datasets into soft labels, which consist of the predicted pseudo-probabilities of every class for each datapoint. We then use the soft labels to train our significantly less complex student models, which are based on the gated recurrent unit (GRU) and multilayer perceptron (MLP). The evaluation results using two real-world animal behavior classification datasets show that the classification accuracy of the student GRU-MLP models improves appreciably through KD, approaching that of the teacher ResNet model. To further reduce the computational and memory requirements of performing inference using the student models trained via KD, we utilize dynamic fixed-point quantization (DQ) through an appropriate modification of the computational graph of the considered models. We implement both unquantized and quantized versions of the developed KD-based models on the embedded systems of our purpose-built collar and ear tag devices to classify animal behavior in situ and in real time. Our evaluations corroborate the effectiveness of KD and DQ in improving the accuracy and efficiency of in-situ animal behavior classification.

\end{abstract}


\end{frontmatter}

\section{Introduction}

Accurate knowledge of animal behavior, which is an important indicator of health, welfare, productivity, and performance, can greatly help with efficient management of livestock or wildlife. Manual observation and recording of behavior for large numbers of animals over long periods is impractical as the associated costs or logistical challenges can be prohibitive. Therefore, classifying animal behavior using sensor data, e.g., inertial measurement data that can be collected via wearable devices such as collar or ear tags, is highly desirable~\citep{ca0}.

Most existing works on animal behavior classification using sensor data, or more broadly times-series classification, can be divided into three categories, i.e., those based on hand-tuned thresholds in conjunctions with some simple statistical features in time or frequency domains, conventional feature-engineering-based approaches that involve separate procedures for crafting and calculating the features and learning the classification model, and end-to-end approaches that calculate the features and learn the classification model jointly.

The approaches within the first category, such as~\citep{busch2017determination, williams2019application}, generally achieve good accuracy when classification of a single behavior is concerned and are usually easy to implement on embedded systems or edge devices. However, such solutions are often unsuitable for multi-class classification due to their compounded complexity and poor performance.

The approaches falling into the second category have conventionally formed the mainstream literature regarding time-series data classification and particularly animal behavior classification from accelerometry data. Some examples are~\citep{arablouei2021situ, bagnall2015time, bagnall2017great, baydogan2013bag, bostrom2015binary, deng2013time, kate2016using, lines2015time, rahman2018cattle, schafer2015boss, smith2016behavior}. These approaches calculate various statistical features in time or frequency domains and feed them into supervised machine learning algorithms such as support vector machine, random forest, or multilayer perceptron. They usually achieve good performance in classifying multiple behaviors. Some of them are designed to be implemented on embedded systems or edge devices for in-situ inference. With these approaches, as the feature calculation and classification parts are optimized separately, the features need be carefully engineered and calculated prior to training the classification model.

In the approaches belonging to the third category, such as~\citep{rahman2016comparison, kasfi2016convolutional, peng2019classification, peng2020dam, wang2021animal, arablouei2021animal}, there are learnable parameters in both feature calculation and classification parts of the underlying end-to-end models. Therefore, the parameters related to feature calculation are learned from data together with the classifier parameters rather than being determined separately. The downside of these approaches is their relatively high complexity as they often rely on large models, which are predominantly deep neural networks, to extract meaningful features directly from data. This can result in large memory requirement and processing time, which may prohibit deployment of such end-to-end models on embedded systems or edge devices with limited memory, computational, or energy resources. 

The notion of knowledge distillation (KD), originally proposed in~\citep{hinton2015distilling}, can help us realize accurate end-to-end classification on embedded systems or edge devices while respecting their resource limitations. With KD, we can utilize small models, which fit in memory and run in reasonable time but may not be highly accurate on their own, and let them embody the intelligence of larger models, which are more accurate but too complex to run on target embedded systems or edge devices. KD, in principle, allows us to improve in-situ behavior classification performance at the expense of more costly off-device training.

The idea behind KD, as introduced in~\citep{hinton2015distilling}, is to transfer knowledge from a large model, called the teacher, to a typically smaller one, called the student, without loss of validity. It involves two major steps that are 1) train the teacher model and, for each datapoint, calculate and record the teacher model's output as predicted pseudo-probabilities corresponding to all classes and 2) train the student model using the pseudo-probabilities predicted by the teacher model (soft labels) as well as the ground truth (hard labels). The soft labels generated by the teacher model encode information about how the teacher model represents the knowledge that it extracts from the training dataset. Therefore, the soft labels help the student model benefit from the concise knowledge representation that is learned by the teacher model but is not attainable by the student model alone due to its insufficient capacity.

\subsection{Contributions}

In this paper, we examine the use of KD for enhancing the accuracy of recurrent neural network (RNN)-based models for classifying animal behavior from accelerometry data. Our evaluations using two cattle behavior classification datasets collected during real-world animal trials show that KD can substantially improve the accuracy of the considered RNN-based student models via exploiting the knowledge distilled by accurate but cumbersome teacher models. In particular, KD can drive the classification accuracy of some student models close to that of their teacher model.

To further accelerate the procedure of inference via the models enhanced through KD, we apply dynamic fixed-point quantization (DQ) to the model parameters as well as the associated matrix-vector multiplications (MVMs). We also substitute the computationally-demanding activation functions with their more efficient approximations.

We implement the unquantized and quantized models on the embedded systems of our cattle collar and ear tags and evaluate their inference complexity. The experimental results indicate that DQ and activation function approximation substantially reduce the memory usage and latency of performing inference using the considered models without incurring any significant loss of accuracy.

We show that the considered RNN-based models, whose accuracy is enhanced via KD and complexity reduced via DQ, run smoothly on the embedded systems of our devices without straining the available memory, computational, or energy resources and hence deliver accurate in-situ and real-time classification of animal behavior. This is notwithstanding that running the large teacher models, which are used to make the considered RNN-based models more accurate through KD, on our devices is not feasible due to their high memory and computational demands.

\section{Overview of proposed approach}

\begin{figure}[!t]
    \centering
    \includegraphics[scale=.275]{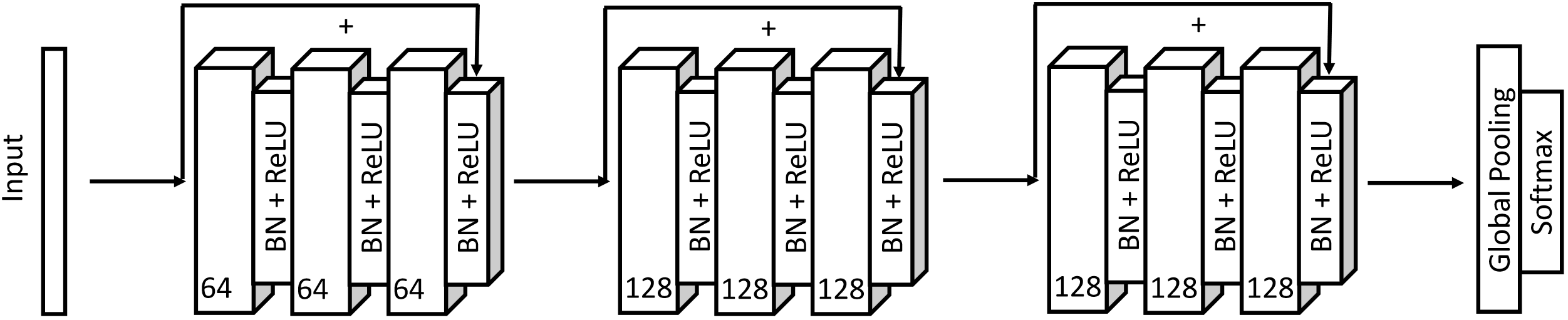}
    \caption{The architecture of the ResNet model~\citep{wang2017time}.}
    \label{fig:resnet}
\end{figure}

We employ KD to improve the accuracy of the RNN-based models proposed in~\citep{wang2021animal} for animal behavior classification. We choose a nine-layer deep residual convolutional neural network, proposed in~\citep{wang2017time} for end-to-end classification of multivariate time-series data, as the teacher network. The architecture of this model, called ResNet, is sketched in Fig.~\ref{fig:resnet}. It is shown in~\citep{fawaz2019deep} that ResNet is one of the most accurate existing time-series classification algorithms, especially among those based on neural networks. 

\begin{figure}[!t]
    \centering
    \includegraphics[scale=.18]{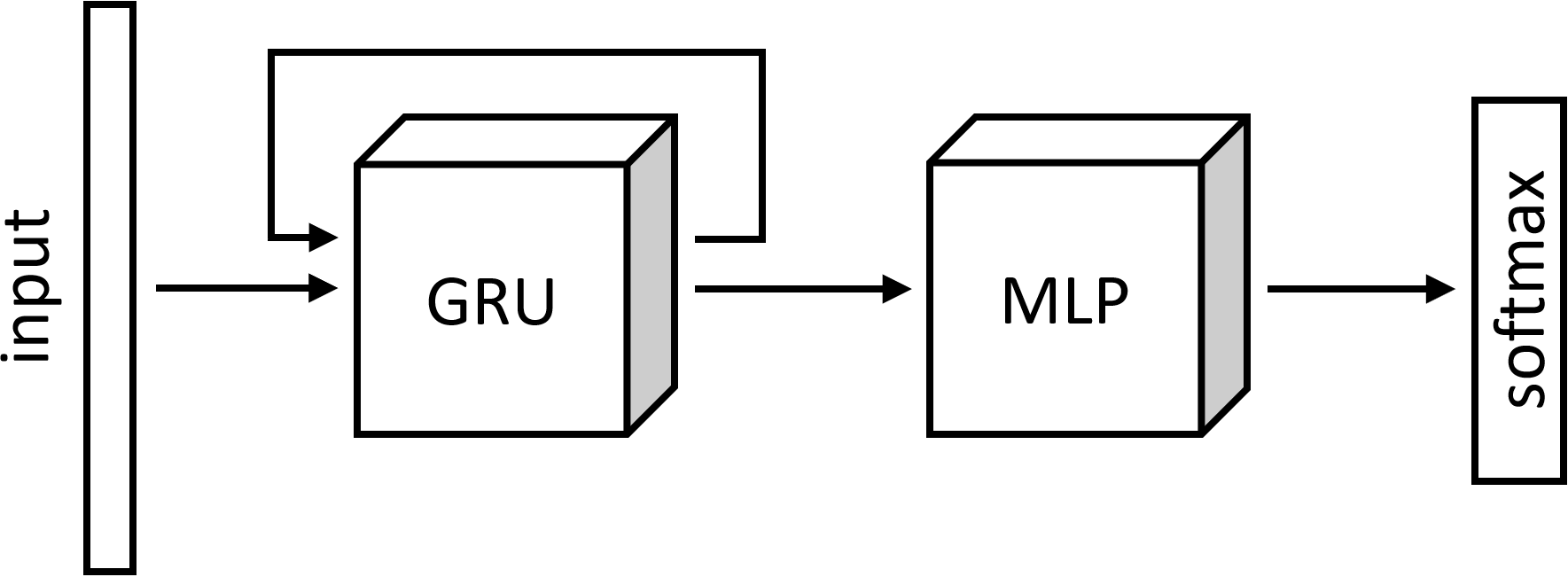}
    \caption{The architecture of the GRU-MLP models proposed in~\citep{wang2021animal}.}
    \label{fig:gru}
\end{figure}

Our student models are composed of one or two layers of the unidirectional gated recurrent unit (GRU) and a multilayer perceptron (MLP) with one hidden layer as proposed in~\citep{wang2021animal}. We sketch the architecture of the student models, which we call GRU-MLP, in Fig.~\ref{fig:gru}. Their computational complexity and memory requirement are significantly smaller compared to the teacher model, ResNet. This is evident in Table~\ref{tab:complex} where we give the number of model parameters of the teacher and student models and the number of multiplication operations required by them for inference. As shown in Table~\ref{tab:complex}, we consider three variants for the student models, which are characterized by the number of GRU layers used in them (1 or 2) and the number of hidden states in each GRU layer (32 or 64). 

\begin{table}[!t] \footnotesize
\caption{The number of parameters of the considered student and teacher models and the number of multiplication operations required by each model to perform three-class classification inference on a single datapoint made up of 256 triaxial accelerometer readings.\\} 
\label{tab:complex}
\centering
\begin{tabular}{l c c c c}
\toprule
model & role & parameters ($\times10^3$) & multiplications ($\times10^6$)\\ 
\midrule
GRU(1,64)-MLP & student & 25.5  & 6.4\\ 
GRU(2,32)-MLP & student & 12.9  & 3.2\\ 
GRU(1,32)-MLP & student & 6.6   & 1.6\\ 
ResNet        & teacher & 520.2 & 132.6\\ 
\bottomrule
\end{tabular}
\end{table}

To improve the efficiency of performing animal behavior inference using the considered GRU-MLP models learned through KD, we apply DQ where we quantize the floating-point 32-bit (FP32) parameter values of the models into fixed-point 8-bit (Q7) numbers and convert the FP32 MVMs to their equivalent Q7 operations. To this end, we alter the computational graph of the GRU-MLP models appropriately.

To further speed up the inference via the developed models, we replace the sigmoid and hyperbolic tangent (tanh) activation functions with their rational expression (fraction of polynomials) approximations obtained via Gauss's continued fraction with seven divisions. The approximate activation functions are significantly more efficient to compute.

We provide the detailed description of the KD and DQ processes, which we utilize in the proposed approach, in sections~\ref{sec:kd} and~\ref{sec:dq}, respectively. We also present the approximation of the sigmoid and tanh functions in section~\ref{sec:apr}.

\section{Knowledge distillation} \label{sec:kd}

In this section, we provide a brief overview of the KD process that we consider in this work and describe how we set the values of the related hyperparameters.

\subsection{Training}

\begin{figure}
    \centering
    \includegraphics[scale=.37]{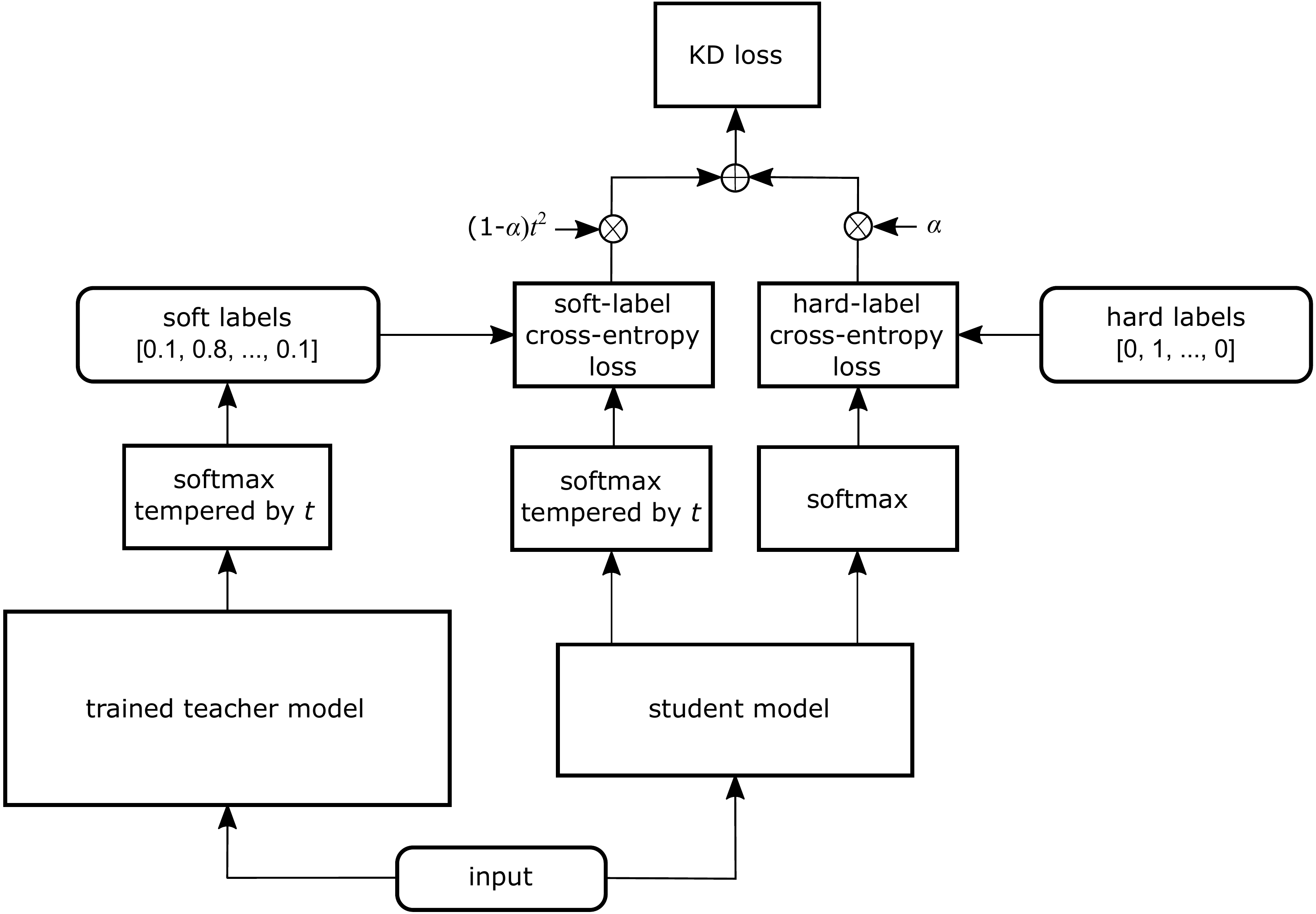}
    \caption{The diagram of calculating the KD loss when training the student model.}
    \label{fig:kd}
\end{figure}

The first step of the KD process is to train the teacher model followed by storing the pseudo-probabilities of all classes predicted by the trained teacher model for each datapoint in the training dataset. The pseudo-probabilities are called soft labels or soft targets and carry information on how confident the trained teacher model is about any datapoint belonging to each class. The soft labels, which can also be viewed as confidence scores, are thought to constitute a concise representation of the knowledge within the dataset that is relevant to the classification task at hand.

The second step is to train the student model using both hard and soft labels. The difference in training of the student model brought about by KD is due to the use of the soft labels produced by the trained teacher model. Hence, the loss function associated with training the student model through KD is defined as
\begin{equation} \label{equ:l_kd}
   l_{k} = \alpha l_{h} + (1-\alpha)t^2 l_{s}
\end{equation}
where $l_h$ and $l_s$ are the loss functions corresponding to the hard and soft labels, respectively, $\alpha$ is the mixing parameter, and $t$ is the temperature parameter, which we will explain in the following. Fig. \ref{fig:kd} depicts the calculation of the KD loss.

For any given datapoint, the hard-label loss function, $l_h$, is defined as the cross entropy of the one-hot vector of the ground-truth label and the corresponding softmax output vector of the student model, i.e.,
\begin{equation}
    l_{h} = - \sum_{i=1}^C s_i \log(q_i) = -\log(q_c)
\end{equation}
where $s_i$ is the $i$th entry of the label one-hot vector, $C$ is the number of classes, $c$ is the index of the ground-truth class label, and $q_i$ is the $i$th entry of the softmax output vector expresses as
\begin{equation}
    q_i = \frac{\exp{(z_i)}}{\sum_{j=1}^C \exp{(z_j)}}
\end{equation}
with $z_i$ being the output logit corresponding to the $i$th class produced by the student model.

The soft-label loss function, $l_s$, is the cross entropy of the soft labels produced by the trained teacher model and the softmax output of the student model, both smoothed by raising the temperature from the default value of $1$ to $t>1$ through the division of the related logits by $t$, i.e.,
\begin{equation}\label{sl}
    l_{s} = - \sum_{i=1}^C p_{i}^{(t)} \log\left(q_{i}^{(t)}\right)
\end{equation}
where
\begin{equation}\label{pq}
    q_{i}^{(t)} = \frac{\exp{\left(\dfrac{z_i}{t}\right)}}{\sum_{j=1}^C \exp{\left(\dfrac{z_j}{t}\right)}},\quad
    p_{i}^{(t)} = \frac{\exp{\left(\dfrac{v_i}{t}\right)}}{\sum_{j=1}^C \exp{\left(\dfrac{v_j}{t}\right)}},
\end{equation}
and $v_i$ is the output logit of the trained teacher model corresponding to the $i$th class.

Since the gradient of $l_s$ has an additional multiplicative factor of $1/t^2$ compared with the gradient of $l_h$, we multiply the second term on the right-hand side of~\eqref{equ:l_kd} by $t^2$ so that the contributions of the gradients of $l_h$ and $l_s$ have similar orders regardless of the value of $t$.

\subsection{Hyperparameters}

Here, we discuss the choice of the values for two important hyperparameters involved in KD, i.e., the temperature $t$ and the mixing parameter $\alpha$.

The mixing parameter $\alpha$ governs the relative contributions of the hard-label and soft-label loss functions. If the value of $\alpha$ is close to zero, the KD loss function is dominated by the soft-label loss function, which means the student model mainly learns to predict pseudo-probabilities similar to those of the teacher model with less emphasis on complying with the ground truth. When the value of $\alpha$ is close to one, the KD loss function is dominated by the hard-label loss function. Hence, the agreement with ground truth is prioritized to learning from the soft labels produced by the teacher model. Since the success of KD is considered to be due to the teacher model's ability to extract insightful information from the training dataset and encode it within the soft labels, a small value for $\alpha$ is preferable. Our experiment results suggest that $\alpha = 0.1$ establishes a good balance between the two loss components.

The value of the temperature $t$ determines the relative contributions of the summand terms on the right-hand side of~\eqref{sl}. Increasing the value of $t$ smooths the distribution of the pseudo-probabilities in~\eqref{pq} and consequently increases their entropy. This allows the pseudo-probabilities with smaller values to have higher relative contribution to the soft-label loss function and therefore training via KD, which has been shown to improve the accuracy of the learned model. On the other hand, if $t$ is too large, the pseudo-probabilities may tend to a uniform distribution and lose their information. In our experiments, $t=3$ appears to deliver the best results in most considered cases.

\section{Quantization and approximation}\label{sec:dq}

Our goal is to accurately classify animal behavior using accelerometry data in situ and in real time. Therefore, in this section, we explain how we utilize DQ and approximate the sigmoid and tanh activation functions to alleviate the burden of executing the considered GRU-MLP models, which are trained via KD, on the embedded systems of our collar and ear tags. In section~\ref{sec:eval}, we show that DQ and the approximations substantially reduce the memory and CPU time required for on-device animal behavior inference with no significant sacrifice of classification accuracy.

\begin{figure}
    \centering
    \includegraphics[scale=.5]{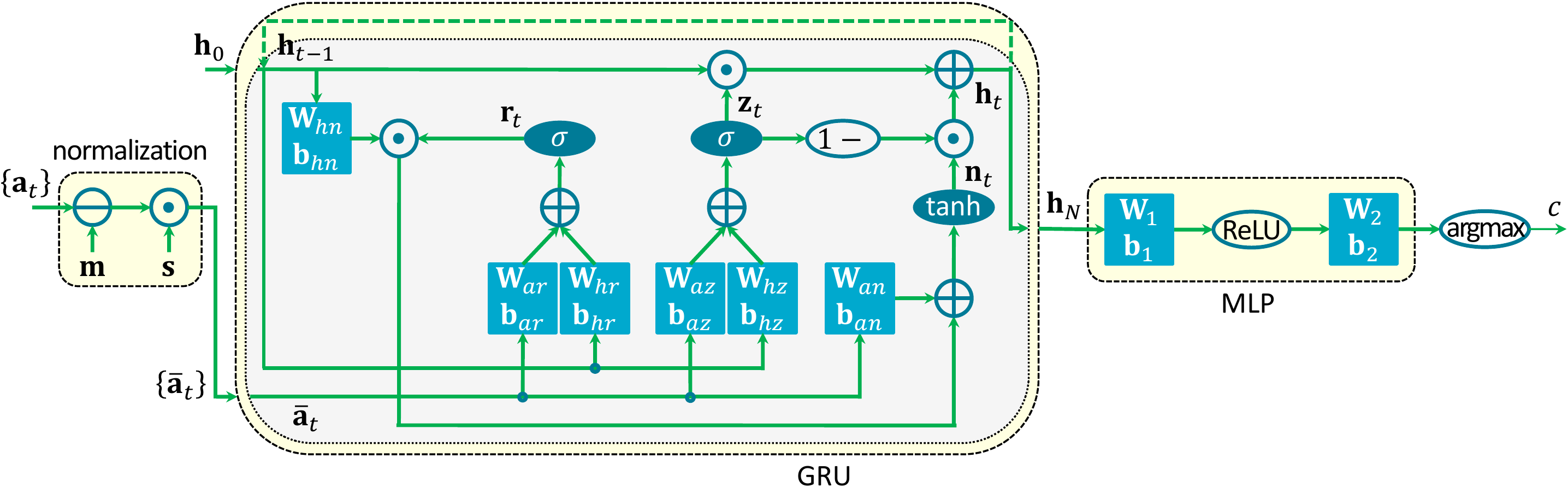}
    \caption{The computational graph of performing inference via a GRU($1$,$L$)-MLP model.}
    \label{fig:cg}
\end{figure}

To make the exposition of our ideas in this section clearer, in Algorithm~\ref{alg}, we summarize the procedure of performing animal behavior classification using a GRU-MLP model with a single GRU layer. In Fig.~\ref{fig:cg}, we also illustrate the computational graph of the inference procedure given in Algorithm~\ref{alg}.

\begin{algorithm}[!t]\label{alg} \footnotesize
\renewcommand{\AlCapSty}[1]{\normalfont\small{#1}\unskip}
\caption{The animal behavior inference procedure using the GRU-MLP model with a single GRU layer including the related parameters.}
\footnotesize
\DontPrintSemicolon
\begin{tabular}{@{}l l}
input:\\
\quad $\mathbf{a}_t\in\mathbb{R}^{3\times 1}, t=1,\dots,N$ & vectors of accelerometer readings in $x$, $y$, and $z$ axes\\
\quad $\mathbf{h}_0\in\mathbb{R}^{L\times 1}$ & initial values of hidden states\\
output:\\
\quad $c\in\{1,\dots,C\}$ & predicted behavior class index\\
parameters:\\
\quad $N\in\mathbb{Z}^+$ & accelerometer reading sequence length\\
\quad $L\in\mathbb{Z}^+$ & number of hidden states\\
\quad $C\in\mathbb{Z}^+$ & number of classes\\
\quad $\mathbf{m}\in\mathbb{R}^{3\times 1}$ & normalization means\\
\quad $\mathbf{s}\in\mathbb{R}^{3\times 1}$ & normalization inverse standard-deviations\\
\quad $\mathbf{W}_{ar},\mathbf{W}_{az},\mathbf{W}_{an}\in\mathbb{R}^{L\times 3}$ & GRU weights\\
\quad $\mathbf{W}_{hr},\mathbf{W}_{hz},\mathbf{W}_{hn} \in\mathbb{R}^{L\times L}$ & GRU weights\\
\quad $\mathbf{b}_{ar},\mathbf{b}_{hr},\mathbf{b}_{az},\mathbf{b}_{hz},\mathbf{b}_{an},\mathbf{b}_{hn} \in\mathbb{R}^{L\times 1}$ & GRU biases\\
\quad $\mathbf{W}_{1}\in\mathbb{R}^{\lfloor(L+C)/2\rfloor\times L},\mathbf{W}_{2}\in\mathbb{R}^{C\times\lfloor(L+C)/2\rfloor}$ & MLP weights\\
\quad $\mathbf{b}_{1}\in\mathbb{R}^{\lfloor(L+C)/2\rfloor\times 1},\mathbf{b}_{2}\in\mathbb{R}^{C\times 1}$ & MLP biases\\
\end{tabular}
\vspace{2pt}\\
algorithm:\\
\quad for $t=1$ to $N$:\\
\quad \quad normalization:
\begin{align*}
\bar{\mathbf{a}}_t=\left(\mathbf{a}_t-\mathbf{m}\right)\odot\mathbf{s}
\end{align*}
\quad \quad uni-directional gated recurrent unit:
\begin{align*}
\mathbf{r}_t&=\sigma\left(\mathbf{W}_{ar}\bar{\mathbf{a}}_t+\mathbf{b}_{ar}+\mathbf{W}_{hr}\mathbf{h}_{t-1}+\mathbf{b}_{hr}\right)\\
\mathbf{z}_t&=\sigma\left(\mathbf{W}_{az}\bar{\mathbf{a}}_t+\mathbf{b}_{az}+\mathbf{W}_{hz}\mathbf{h}_{t-1}+\mathbf{b}_{hz}\right)\\
\mathbf{n}_t&=\tanh\left(\mathbf{W}_{an}\bar{\mathbf{a}}_t+\mathbf{b}_{an}+\mathbf{r}_t\odot\left(\mathbf{W}_{hn}\mathbf{h}_{t-1}+\mathbf{b}_{hn}\right)\right)\\
\mathbf{h}_t&=\left(1-\mathbf{z}_t\right)\odot\mathbf{n}_t+\mathbf{z}_t\odot\mathbf{h}_{t-1}  
\end{align*}
\quad \indent multilayer perceptron:
\begin{align*}
c&=\arg\max\left(\mathbf{W}_{2}\max\left(\mathbf{0},\mathbf{W}_{1}\mathbf{h}_N+\mathbf{b}_{1}\right)+\mathbf{b}_{2}\right)
\end{align*}
\end{algorithm}

It is clear from Algorithm~\ref{alg} and Fig.~\ref{fig:cg} that the most computationally demanding parts of the GRU-MLP model are the affine transformations leading to the calculation of $\textbf{r}_t$, $\textbf{z}_t$, and $\textbf{n}_t$ together with the computation of the sigmoid and tanh activation functions. 

The affine transformations represented by the blue rectangles in Fig.~\ref{fig:cg} involve the multiplication of the input or hidden state vectors by the corresponding parameter (weight) matrixes. The calculation of the MLP output also requires two affine transformations or MVMs. However, the six MVMs carried out in each iteration of every GRU layer are particularly taxing as the number of iterations is equal to the length of the input data sequence, which is $256$ in our case. In addition, the weight matrixes may occupy sizable memory space when stored in full precision. The same arguments regarding the major factors of computational and memory complexity also apply to the GRU-MLP models with more than one GRU layer.

\subsection{Dynamic fixed-point quantization}

To reduce the computational complexity of the MVMs and the memory space occupied by the weight matrixes of the GRU-MLP models, we utilize a DQ scheme where we quantize the weights to 8-bit integers and perform the MVMs via Q7 arithmetic. The main advantage of using fixed-point arithmetic operations is that they reduce the likelihood of overflow and consequent incorrect results. However, this comes at the expense of reduced precision, which is a price worth paying considering that the repercussions of overflow are often more severe compared to precision loss.

We quantize the entries of all model parameter (weight) matrixes, e.g., $\textbf{W}_{ar}$ and $\textbf{W}_{hr}$ in Algorithm~\ref{alg}, which originally have FP32 values, to 8-bit integer values ranging between $-128$ and $127$. To quantize each weight matrix, we use a center-point, also known as zero-point, of zero and calculate the corresponding scale parameter such that all entries of the matrix have quantized values between $-128$ and $127$. Therefore, as an example, for $\textbf{W}_{ar}$, we set the scale parameter to $s_{ar}=\frac{2}{2^8-1}\max\left(|\textbf{W}_{ar}|\right)$ where the absolute value operator, $|\cdot|$, and the maximum value operator, $\max(\cdot)$, apply entry-wise and globally, respectively. Therefore, we calculate the quantized integer values as $\textbf{W}_{arQ}=\left\lfloor s_{ar}^{-1}\textbf{W}_{ar}\right\rceil$ where $\left\lfloor\cdot\right\rceil$ denotes rounding to the nearest integer.

\begin{figure}
    \centering
    \subfigure[unquantized]{\includegraphics[scale=.5]{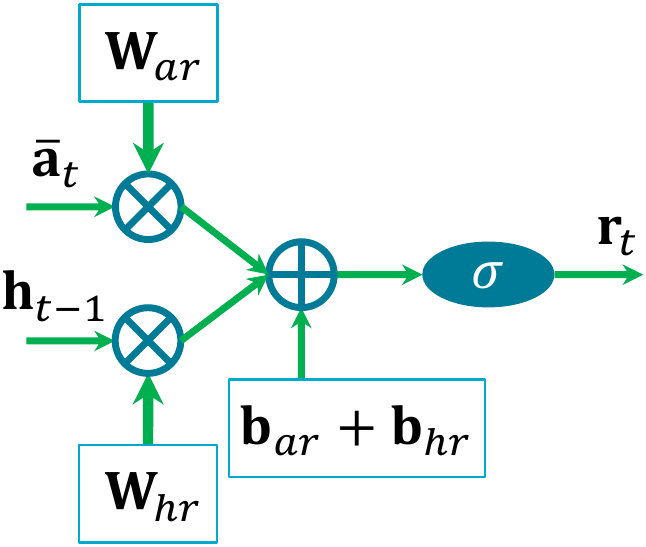}\label{fig:unquant}}
    \hspace{1cm}
    \subfigure[quantized]{\includegraphics[scale=.5]{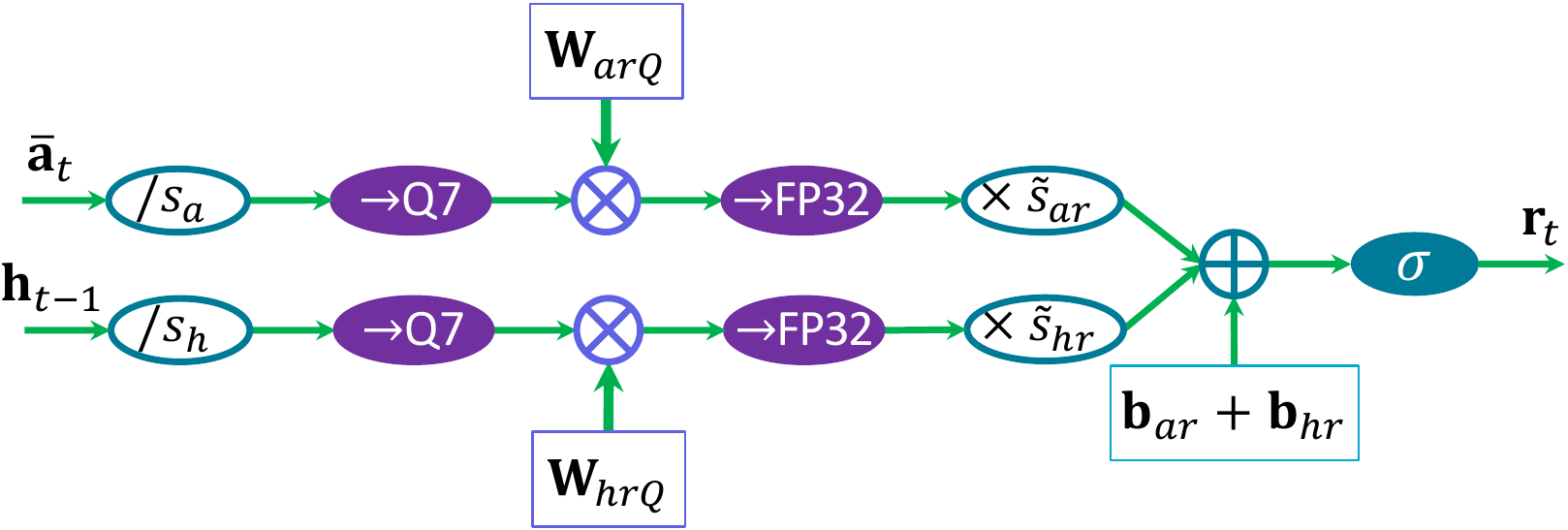}\label{fig:quant}}
    \caption{The computational graphs of calculating $\textbf{r}_t$ with and without DQ.}
    \label{fig:comp_graphs}
\end{figure}

In Fig.~\ref{fig:unquant}, we depict the computational graph of calculating $\textbf{r}_t$ without any quantization. In Fig.~\ref{fig:quant}, we show the modified version of this computational graph where we use the quantized integer weight matrixes, i.e., $\textbf{W}_{arQ}$ and $\textbf{W}_{hrQ}$, instead of the FP32 matrixes $\textbf{W}_{ar}$ and $\textbf{W}_{hr}$ and replace the FP32 MVMs with the corresponding Q7 operations. To utilize Q7 MVMs effectively, we scale the values of the input matrix and input vector appropriately so that the entries of the vector resulting from the multiplication are between $-1$ and $+1$, hence the output vector can be correctly represented by Q7 numbers. Thus, we divide the values of all entries of the vectors $\bar{\textbf{a}}_t$ and $\textbf{h}_{t-1}$ by the input scale parameters denoted by $s_a$ and $s_h$, respectively, to ensure that they are scaled properly before converting them to Q7 format and feeding into Q7 matrix-vector multipliers. After the multiplication, we convert the results back to the FP32 format and rescale them via multiplying by the appropriate rescaling parameters, i.e., $\tilde{s}_{ar}=2^7s_{a}s_{ar}$ and $\tilde{s}_{hr}=2^7s_{h}s_{hr}$ as shown in Fig.~\ref{fig:quant}. Multiplication by $2^7$ is to compensate for the implicit division by $2^7$ that occurs when inputting the 8-bit-integer-valued matrixes, such as $\textbf{W}_{arQ}$ and $\textbf{W}_{hrQ}$, into the respective Q7 MVMs. The implicit division is due to that the Q7 MVM function presumes the entries of the quantized matrixes to be in Q7 format.

We use the same procedure to quantize the weight matrixes and reduce the complexity of the MVMs involved in the calculation of $\textbf{z}_t$ and $\textbf{n}_t$ for every GRU layer and the calculation of the MLP output. We tune the input scale parameters $s_a$ and $s_h$ empirically to achieve the best inference results. To minimize the associated computational burden, we only consider powers-of-two values for these parameters when tuning them.

With DQ, we only store the quantized weight matrixes, which take up four times less memory compared to the original unquantized weight matrixes. Therefore, quantization can provide significant savings in memory usage, even though it incurs a small memory overhead for storing the related scale parameters. Quantization can also reduce inference latency since fixed-point arithmetic operations are generally more efficient than floating-point operations.

\subsection{Approximation of activation functions}\label{sec:apr}

Precise calculation of the sigmoid and tanh functions on embedded systems can be costly with little or no benefit in the context of machine-learning inference using neural networks. Therefore, we implement an approximate version of the tanh function using Gauss's continued fraction with seven divisions~\citep{gcf} that is expressed as
\begin{align}
    \tanh{(x)} \approx& \frac{x}{1+\frac{x^2}{3+\frac{x^2}{5+\frac{x^2}{7+\frac{x^2}{9+\frac{x^2}{11+\frac{x^2}{13}}}}}}}\\
               \approx& \frac{x^7+378x^5+17325x^3+135135x}{28x^6+3150x^4+62370x^2+135135}
\end{align}
and is valid for $x\in[-4.972,4.972]$. For any input $x$ outside this range, we clip the corresponding approximate tanh function value to $-1$ or $+1$ depending on whether $x$ is negative or positive. Accordingly, we approximate the sigmoid function as
\begin{align}
\sigma(x) =& \frac{\tanh{(x)}+1}{2}\\
          \approx& \frac{1}{2}\left(\frac{x^7+378x^5+17325x^3+135135x}{28x^6+3150x^4+62370x^2+135135}+1\right)
\end{align}
and clip it to $0$ or $+1$ when the input is outside the above-mentioned range.

\section{Evaluation}\label{sec:eval}

In this section, we evaluate the classification accuracy and inference complexity of the considered GRU-MLP models that are trained via KD (with the ResNet model as the teacher) and quantized via DQ.

\subsection{Datasets}

In our evaluations, we use two real-world datasets containing labeled accelerometry data collected during an experiment with eight grazing beef cattle. The experiment was conducted in March 2020 at the Commonwealth Scientific and Industrial Research Organisation (CSIRO) FD McMaster Laboratory Pasture In-take Facility~\citep{greenwood2014new}, Chiswick NSW, Australia (\ang{30; 36; 28.17}S, \ang{151; 32; 39.12}E). The experiments were approved by the CSIRO FD McMaster Laboratory Chiswick Animal Ethics Committee with the animal research authority numbers 19/18.

We have collected triaxial accelerometry data using custom-build collar tags\footnote{\url{https://www.csiro.au/en/research/animals/livestock/egrazor-measuring-cattle-pasture-intake}} and ear tags\footnote{\url{https://www.cerestag.com/}} worn by the cattle. The sample rates of the accelerometers on the collar and ear tags were $50$ Hz and $62.5$ Hz, respectively. Therefore, each $256$ consecutive accelerometer readings correspond to time windows of $5.12$s and $4.1$s for the collar and ear tags, respectively.

We have annotated parts of the collected raw accelerometry data by examining the videos of the cattle recorded during the experiment. In this work, consistent with our previous related work in~\citep{wang2021animal}, we consider three behavior classes of grazing, resting, and alia. The resting behavior class also includes the ruminating behavior. The alia class includes all behaviors other than grazing and resting. To avoid confusion, we use the Latin word \textit{alia} instead of \textit{other} to describe the class that encompasses all remaining behaviors. We refer to the datasets corresponding to the data collected via collar tags and ear tags as Arm20c and Arm20e, respectively. In both datasets, each datapoint contains $256$ consecutive triaxial accelerometer readings and their associated behavior class label.

More detailed information about the experiment and the datasets as well as the hardware and software utilized for data collection can be found in~\citep{arablouei2021animal, wang2021animal, acc+gps}.


\subsection{Overall accuracy}

We use the Matthews correlation coefficient (MCC)~\citep{matthews1975comparison} as our classification accuracy metric. The MCC is a suitable accuracy measure when the underlying dataset is imbalanced as it takes into account true and false positives and negatives. Its value is between $-1$ and $+1$ with $+1$ being perfect prediction, $0$ no better than random prediction, and $-1$ perfect inverse prediction.

In addition, we use a leave-one-animal-out cross-validation scheme where, in each fold, we use the data of one animal for validation and the remaining data for training and tuning the relevant hyperparameters. To calculate the cross-validated MCC values, we aggregate the classification results evaluated on the validation data of all folds.

We set the values of the hyperparameters of the ResNet model according to Table 1 of~\citep{fawaz2019deep}, which are shown to be optimal in different applications with various datasets.

\begin{figure}[!t]
    \centering
    \subfigure[The Arm20c dataset.]{\includegraphics[scale=.5]{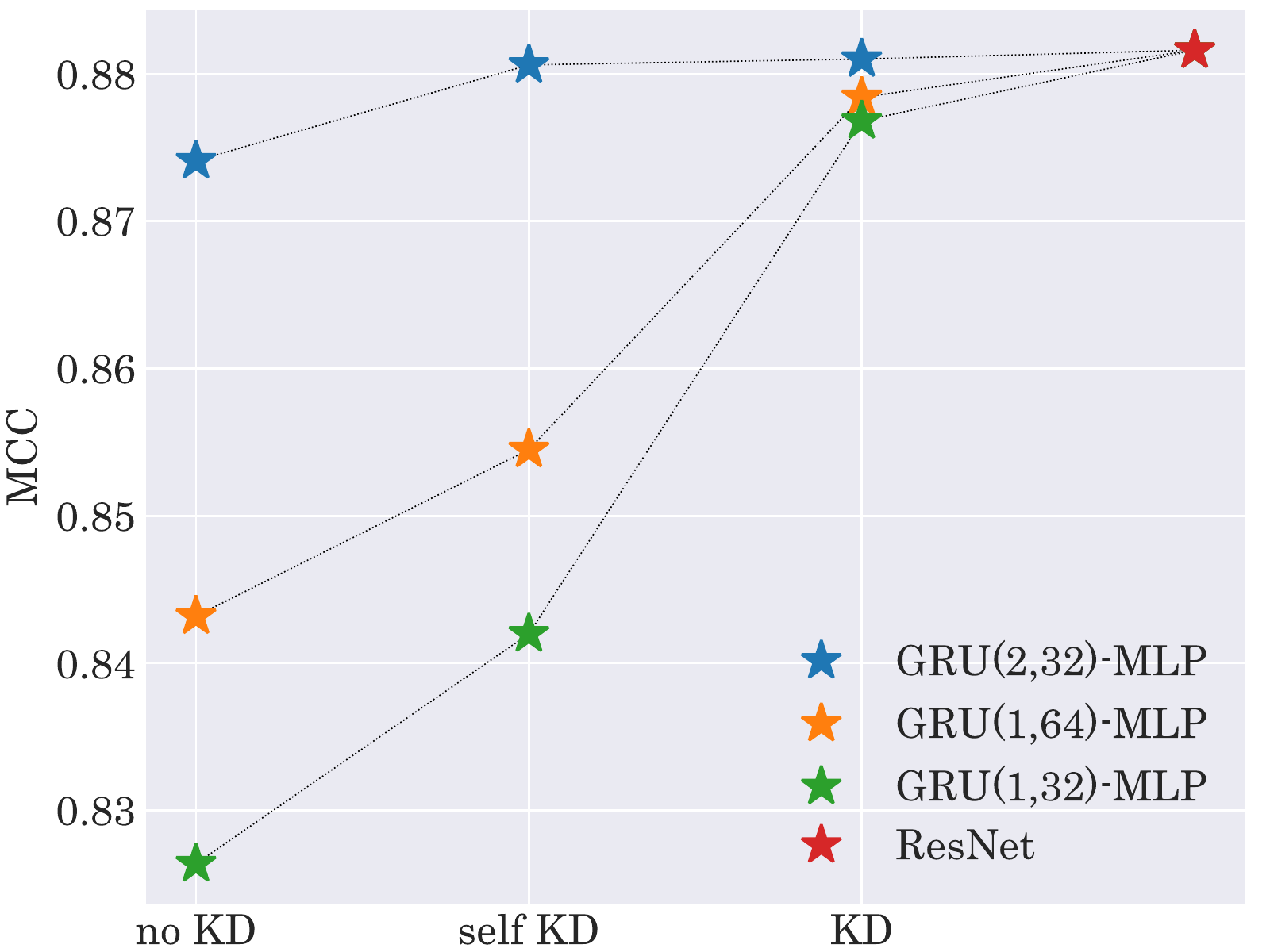}\label{fig:arm20c}}
    \subfigure[The Arm20e dataset.]{\includegraphics[scale=.5]{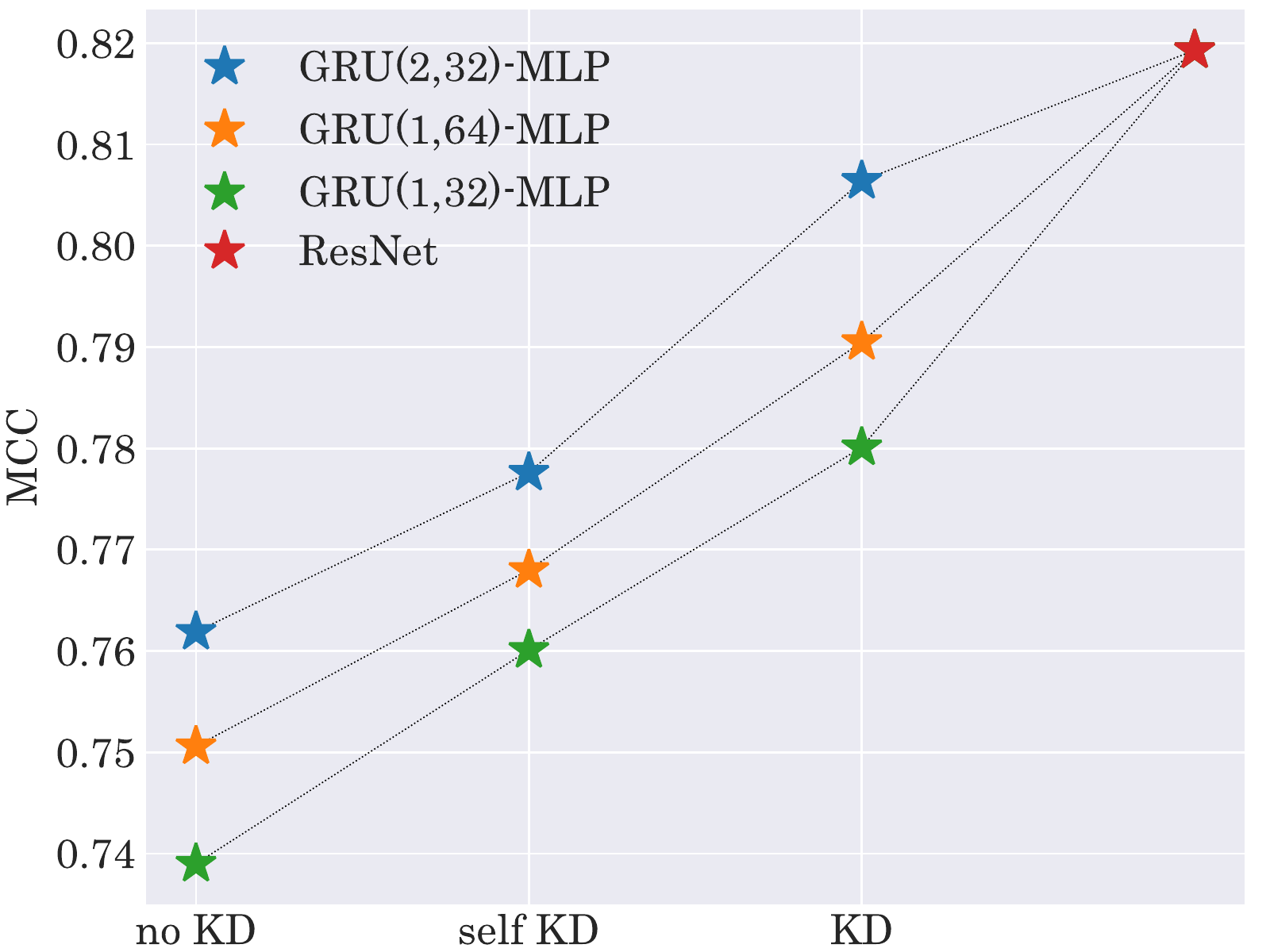}\label{fig:arm20e}}
    \caption{The cross-validated MCC values of the considered student models with and without KD and those of the teacher ResNet model for both considered datasets.}
    \label{fig:mccplots}
\end{figure}

In Table~\ref{tab:mccc}, we present the cross-validated MCC values for the considered student GRU-MLP models when trained with and without KD and those of the teacher ResNet model for both considered datasets. The table contains the multiclass MCC values calculated by taking into account all entries of the respective confusion matrixes~\citep{Rk}. The results for the models trained without any KD are in the \textit{no KD} column and those for the models trained via KD are in the \textit{KD} column. The \textit{self KD} column contains the results for the models that are trained via KD while each model is its own teacher. To facilitate comparison, in Fig.~\ref{fig:mccplots}, we provide a visual representation of the MCC values given in Table~\ref{tab:mccc}. 

\begin{table}[!t] \footnotesize
\caption{The cross-validated MCC values of the considered student models with and without KD and those of the teacher ResNet model for both considered datasets.\\}
\label{tab:mccc}
\centering
\begin{tabular}{l l c c c}
\toprule
dataset & model        & no KD & self KD & KD \\ 
\midrule
Arm20c & GRU(2,32)-MLP & 0.874 & 0.881 & 0.881\\ 
& GRU(1,64)-MLP        & 0.843 & 0.855 & 0.878\\
& GRU(1,32)-MLP        & 0.826 & 0.842 & 0.877\\ 
& ResNet               & 0.882 &&\\
\midrule
Arm20e & GRU(2,32)-MLP & 0.762 & 0.778 & 0.806\\
& GRU(1,64)-MLP        & 0.751 & 0.768 & 0.791\\
& GRU(1,32)-MLP        & 0.739 & 0.760 & 0.780\\
& ResNet               & 0.819 &&\\
\bottomrule
\end{tabular}
\end{table}

The results in Table~\ref{tab:mccc} and Fig.~\ref{fig:mccplots} demonstrate that KD appreciably improves the classification accuracy for all considered models and datasets. The improvement due to KD with the ResNet model is significant when evaluated over both datasets. However, it is more pronounced with the Arm20c dataset, where KD drives the MCC values of all three GRU-MLP models close to that of the teacher ResNet model. The accuracy improvement offered by KD with the Arm20e dataset appears to be not as noticeable as that with the Arm20c dataset. However, KD is still significantly helpful with the Arm20e dataset as classifying animal behavior using ear tag accelerometry data is inherently more challenging compared to using collar tag data~\citep{wang2021animal, acc+gps}.  

As expected, KD using the ResNet model as the teacher is more effective compared with using the GRU-MLP models as their own teachers. However, the results suggest that there still exists substantial benefits to KD, even when the GRU-MLP models teach themselves. This attests to the efficacy of KD in providing a more practical classification target through the use of soft labels, even if they do not come from a more accurate teacher model. In addition, self KD appears to be more effective when the models are learned from the Arm20e dataset. This may also be attributable to the comparatively higher complexity of recognizing animal behavior from ear tag accelerometry data.

\subsection{Accuracy for each behavior}

In Table~\ref{tab:mcc_pb}, we provide the cross-validated MCC values corresponding to each considered behavior class for all considered models with or without KD and for both considered datasets. The per-class MCC values are calculated by treating the classification of each behavior class as a binary classification problem.

\begin{table}[!t] \footnotesize
\caption{The cross-validated MCC values corresponding to each considered behavior class for all considered models with and without KD and for both considered datasets.\\}
\label{tab:mcc_pb}
\centering
\begin{tabular}{l l l c c c}
\toprule
dataset & behavior & model         & no KD & self KD & KD\\ 
\midrule
Arm20c & grazing   & GRU(2,32)-MLP & 0.908 & 0.909 & 0.909\\
       &           & GRU(1,64)-MLP & 0.898 & 0.906 & 0.909\\
       &           & GRU(1,32)-MLP & 0.884 & 0.898 & 0.908\\
       &           & ResNet        & 0.909 &&\\
\cline{2-6}
       & resting   & GRU(2,32)-MLP & 0.908 & 0.911 & 0.918\\
       &           & GRU(1,64)-MLP & 0.874 & 0.884 & 0.914\\  
       &           & GRU(1,32)-MLP & 0.868 & 0.877 & 0.912\\
       &           & ResNet        & 0.920 &&\\
\cline{2-6}
       & alia     & GRU(2,32)-MLP & 0.750 & 0.755 & 0.765\\  
       &           & GRU(1,64)-MLP & 0.700 & 0.718 & 0.763\\
       &           & GRU(1,32)-MLP & 0.666 & 0.688 & 0.757\\
       &           & ResNet        & 0.766 &&\\
\hline
Arm20e & grazing   & GRU(2,32)-MLP & 0.851 & 0.862 & 0.880\\
       &           & GRU(1,64)-MLP & 0.834 & 0.856 & 0.872\\
       &           & GRU(1,32)-MLP & 0.835 & 0.853 & 0.870\\
       &           & ResNet        & 0.895 &&\\
\cline{2-6}
       & resting   & GRU(2,32)-MLP & 0.786 & 0.810 & 0.831\\
       &           & GRU(1,64)-MLP & 0.784 & 0.807 & 0.819\\  
       &           & GRU(1,32)-MLP & 0.782 & 0.804 & 0.812\\
       &           & ResNet        & 0.839 &&\\
\cline{2-6}
       & alia      & GRU(2,32)-MLP & 0.534 & 0.537 & 0.604\\
       &           & GRU(1,64)-MLP & 0.539 & 0.543 & 0.577\\
       &           & GRU(1,32)-MLP & 0.500 & 0.522 & 0.558\\
       &           & ResNet        & 0.638 &&\\
\bottomrule
\end{tabular}
\end{table}

Grazing is the most common behavior observed in grazing beef cattle. It is also the most prevalent behavior in our datasets, which represent the real life rather well. The grazing behavior leaves a distinct signature in triaxial accelerometry data, particularly the data collected by collar tags. This is mainly because during grazing the animal lowers its head and moves it relatively vigorously. Therefore, the considered GRU-MLP models can classify the grazing behavior with good accuracy even without KD. Nevertheless, KD improves the accuracy of classifying the grazing behavior, especially with the Arm20e dataset and the models that have a single GRU layer, i.e., GRU(1,32)-MLP and GRU(1,64)-MLP. Using the Arm20c dataset and the ResNet model as the teacher, the per-class MCC values corresponding to the grazing behavior approach that of the ResNet model for all considered student models trained via KD.

Classification of the resting and alia behavior classes is more challenging compared to the grazing behavior. The resting behavior, which here includes the ruminating behavior, may occasionally be confused with other behaviors that occur with low levels of animal body motion, such as drinking or searching, which are considered to fall into the alia behavior class.

Overall, the results of Table~\ref{tab:mcc_pb} show that KD can significantly improve the classification accuracy of all considered behavior classes with both collar and ear tag accelerometry data.

\subsection{Accuracy of quantized models}

Here, we evaluate the accuracy of classifying animal behavior via the quantized models over both Arm20c and Arm20e datasets and present the resulting cross-validated MCC values in Table~\ref{tab:mccq} together with the MCC values for the unquantized models. In Table~\ref{tab:mccq}, we also give the difference between the cross-validated MCC values of the unquantized and quantized versions of each model. As seen in Table~\ref{tab:mccq}, the difference between the MCC values of the unquantized and quantized versions of all models is small. Therefore, the loss of classification accuracy due to dynamic fix-point quantization is practically negligible for all considered GRU-MLP models.

\begin{table}[!t] \footnotesize
\caption{The cross-validated MCC values for both unquantized and quantized versions of the considered models trained via KD with ResNet as the teacher and for both considered datasets, and the difference between the MCC values of the unquantized and quantized versions for each model.\\}
\label{tab:mccq}
\centering
\begin{tabular}{l l c c c}
\toprule
dataset & model        & unquantized & quantized & difference\\ 
\midrule
Arm20c & GRU(2,32)-MLP & 0.881       & 0.874     & 0.007\\ 
& GRU(1,64)-MLP        & 0.855       & 0.851     & 0.004\\ 
& GRU(1,32)-MLP        & 0.842       & 0.839     & 0.003\\ 
\midrule
Arm20e & GRU(2,32)-MLP & 0.778       & 0.772     & 0.006\\ 
& GRU(1,64)-MLP        & 0.768       & 0.765     & 0.003\\ 
& GRU(1,32)-MLP        & 0.760       & 0.758     & 0.002\\ 
\bottomrule
\end{tabular}
\end{table}

\subsection{Complexity of in-situ inference}

We implement performing inference using both unquantized and quantized versions of the considered student models containing one or two GRU layers on the embedded systems of our bespoke cattle collar and ear tags and evaluate their memory usage and on-device runtime.

We implement the considered GRU-MLP models trained via KD on the embedded systems of our collar and ear tags for inferring animal behavior in situ and in real time. To evaluate the savings afforded by DQ in terms of memory usage and CPU runtime, we implement both unquantized and quantized versions of each model. Our target embedded system utilizes a Nordic nRF52840\footnote{\url{https://www.nordicsemi.com/products/nrf52840}} system-on-chip microcontroller containing a 64MHz ARM Cortex-M4F CPU, 256kB of random-access memory (RAM), 1MB of flash read-only memory (ROM), and a floating-point unit (FPU).

In our implementations, we use the Arm Common Microcontroller Software Interface Standard (CMSIS)’s DSP and NN libraries\footnote{\url{https://developer.arm.com/tools-and-software/embedded/cmsis}} and the Zephyr real-time operation system\footnote{\url{https://www.zephyrproject.org/}}. CMSIS is a vendor-independent abstraction layer for microcontrollers that are based on Arm Cortex processors.

In Table~\ref{tab:impl_comp}, we provide the required memory and CPU runtime of performing inference on a single datapoint, i.e., $256$ consecutive triaxial accelerometer readings, using both unquantized and quantized versions of the considered GRU-MLP models. The terms \enquote{text} and \enquote{rodata} refer to the ROM space occupied by the algorithm code and the model parameters, respectively, while \enquote{stack} refers to the RAM space required to store all variables when running each model.

\begin{table}[!t] \footnotesize
\caption{The complexity of performing on-device inference using the consider GRU-MLP models with and without quantization in terms of CPU time and memory usage.\\}
\label{tab:impl_comp}
\centering
\begin{tabular}{l l c c c c}
\toprule
            & model              & CPU time (ms) & rodata (kB) & text (kB) & stack (kB)\\
\midrule
unquantized & GRU(2,32)-MLP & 723          & 41.80       & 3.28      & 2.85\\
            & GRU(1,64)-MLP & 892          & 62.02       & 2.76      & 2.91\\
            & GRU(1,32)-MLP & 286          & 16.71       & 2.76      & 2.26\\
\midrule
quantized   & GRU(2,32)-MLP & 314          & 11.29       & 4.23      & 3.47\\
            & GRU(1,64)-MLP & 371          & 16.39       & 3.55      & 4.35\\
            & GRU(1,32)-MLP & 139          & 4.63        & 3.55      & 2.33\\
\bottomrule
\end{tabular}
\end{table}

The results in Table~\ref{tab:impl_comp} show that the runtime of each quantized model is less than half of that of its unquantized counterpart. In addition, although quantization slightly increase the RAM usage, it yields considerable reduction in ROM usage. 

During a field trial with eight Angus beef cows in February 2022, we performed in-situ behavior classification on the embedded systems of our collar and ear tags using both unquantized and quantized versions of the considered GRU-MLP models, which were trained via KD over the Arm18 and Arm20 datasets. All models ran effectively in real time inferring cattle behavior with classification accuracy on par with the results presented in Table~\ref{tab:mccc}.

\section{Concluding remarks}\label{sec:discussion}

Using end-to-end deep leaning models such as ResNet, animal behavior can be classified accurately from accelerometry data collected by wearable sensors. Deep neural networks have high learning ability owing to their aptly-crafted deep and complex architectures. However, they are large and resource-hungry hence not suitable for implementation on embedded systems or edge devices, which have limited computational, memory, and energy resources. The notion of KD allows the knowledge gained by a complex and accurate model to be transferred to a less complex one that demands less resources and can be implemented on embedded systems or edge devices for in situ and real time inference. As a result, KD enhances the accuracy of classification models that are suitable for on-device inference by utilizing the knowledge acquired from training datasets through more accurate and complex teacher models. KD enables more accurate inference via low-complexity models at the cost of more laborious training. Therefore, KD essentially allows trading training complexity for inference accuracy and simplicity.

Our findings corroborate the merits of KD, even in its most fundamental form, i.e., utilizing soft labels. It enables accurate on-device classification of animal behavior using accelerometry time-series data through imparting the knowledge of a complex model such as ResNet to less complex ones such as GRU-MLP by means of the soft labels generated by the ResNet model. KD is effective even when a GRU-MLP model is used as its own teacher. This points to the fundamental shortcomings of the cross-entropy loss with hard labels in training classification models. When a classification model has restricted learning capacity and the amount of training data is limited, estimating the model's parameters through optimizing the cross-entropy loss with hard labels may become a highly underdetermined problem. Having less strict and more nuanced soft labels as the classification target instead of the hard labels may alleviate this problem thanks to the additional information afforded by the soft labels. This is possible even when the soft labels are produced by the student model itself and not a more accurate teacher model.

We use two real-world datasets collected via cattle collar and ear tags to validate the advantages of KD and quantization for on-device animal behavior classification using accelerometry data. The two datasets have substantial dissimilarities since the utilized collar and ear tags are worn by the cattle differently, i.e., on top of the animal's neck versus on its ear, and have different accelerometer chips with different sampling rates. Thus, the considered models perform considerably differently when evaluated using different datasets. This also applies to the accuracy improvement due to KD. However, regardless of the dataset used, the advantages of KD are evident, albeit to varying degrees. In addition, it appears that, the more challenging a particular behavior is to classify correctly, the more KD improves its classification accuracy.

Our evaluations attest to the effectiveness of DQ in reducing the inference time and memory requirement of the considered GRU-MLP models. The benefits of quantization come with no significant loss in classification accuracy. The savings in inference latency and memory usage offered by quantization are critical when the inference is implemented on resource-constrained devices involving embedded systems or edge devices. DQ provides significant reduction in runtime by cutting it almost in half, despite the utilized microcontroller being equipped with an FPU. This can be attributed to the fact that 8-bit DQ allows Cortex-M4 CPU to utilize the single instruction multiple data (SIMD) feature since Q7 arithmetic operations are performed on 8-bit (one-byte) numbers while FP32 operations are performed on 32-bit (4-byte) numbers. SIMD enables multiple 8-bit or 16-bit operations in a single cycle, i.e., simultaneous computation of two 16-bit or four 8-bit operands, with a near zero increase in power consumption~\citep{cortex}.

It is noteworthy that, despite leading to a significant decrease in the precision of the model parameters and the related MVMs, DQ has little effect on the classification accuracy of the considered GRU-MLP models. The GRU entails a recursive procedure that may result in propagation and accumulation of quantization errors. However, the quantization errors accumulate only to a limited extent over the initial iterations, after which the associated cumulative error plateaus. This is likely because the quantization errors are distributed symmetrically around zero hence cancel out statistically when added together.

In this work, we only considered dynamic quantization where only the model weight matrixes are quantized. Dynamic quantization is favorable when loading the weight matrixes from the memory constitutes a significant portion of the model execution time. It is generally the preferred method to quantize recurrent neural network such as GRU. In future work, we will examine the use of static quantization where both model parameters and activations are quantized. We will also consider quantization-aware training by modeling the effects of quantization during training.

\section*{Acknowledgments}

This research was undertaken with strategic investment funding from the CSIRO and NSW Department of Primary Industries. We would like to thank the technical staff who were involved in the research at CSIRO FD McMaster Laboratory Chiswick, i.e., Flavio Alvarenga, Alistair Donaldson, and Reg Woodgate of the NSW Department of Primary Industries, and Jody McNally and Troy Kalinowski of the CSIRO Agriculture and Food. We also recognize the contributions of the CSIRO Data61 staff who have designed and built the hardware and software of the devices used for data collection, specifically, John Scolaro, Leslie Overs, and Stephen Brosnan.

\bibliographystyle{elsarticle-harv} 
\bibliography{biblio}

\end{document}